\documentclass{iosart2c}
\pdfoutput=1
\usepackage[T1]{fontenc}
\usepackage{times}%
\usepackage{natbib}
\usepackage{amsmath}
\usepackage{amsfonts}
\usepackage{graphics}
\usepackage{epstopdf}
\usepackage{dcolumn}
\usepackage{graphics}
\usepackage{tikz}
\usepackage{pgfplots}
\usepackage{sansmath}
\usepackage{pgfplotstable}
\usepackage{placeins}
\usepackage{booktabs}
\usepackage{enumerate}

\pgfplotsset{compat=newest} 
\usepgfplotslibrary{groupplots}
\usetikzlibrary{shapes}
\pgfplotsset{scaled y ticks=false,
  tick label style = {font=\sansmath\sffamily\scriptsize},
  every axis label = {font=\sansmath\sffamily\scriptsize},
  legend style = {font=\sansmath\sffamily\footnotesize},
  label style = {font=\sansmath\sffamily\scriptsize},
  title style = {font=\sansmath\sffamily\footnotesize},
}

\usepackage{url}
\usepackage{comment}
\newcolumntype{d}[1]{D{.}{.}{#1}}

\begin{document}
\begin{frontmatter}                     
\title{Improving unsupervised neural \\ aspect extraction for online discussions using~out-of-domain classification}
\runningtitle{Improving unsupervised aspect extraction using~out-of-domain classification}
\runningauthor{A. Alekseev et al.}

\author[A]{\fnms{Anton} \snm{Alekseev}\thanks{Corresponding author. E-mail:~anton.m.alexeyev@gmail.com \\ The final publication is available at IOS Press through http://dx.doi.org/10.3233/JIFS-179908. }}
\author[A,C]{\fnms{Elena} \snm{Tutubalina}}
\author[D]{\fnms{Valentin} \snm{Malykh}}
\author[B,A]{\fnms{Sergey} \snm{Nikolenko}}
\address[A]{Samsung-PDMI Joint AI Center, Steklov Mathematical Institute at St.~Petersburg\\191023, 27 Fontanka, St. Petersburg, Russia}
\address[B]{National Research University Higher School of Economics, St. Petersburg, Russia}
\address[C]{Kazan Federal University,
420008, 18 Kremlyovskaya Street, Kazan, Russia}
\address[D]{Moscow Institute of Physics and Technology,
141701, 9 Institutskiy per., Dolgoprudny, Moscow Region, Russia}

\begin{abstract}
Deep learning architectures based on self-attention have recently achieved and surpassed state of the art results in the task of unsupervised aspect extraction and topic modeling. While models such as neural attention-based aspect extraction (ABAE) have been successfully applied to user-generated texts, they are less coherent when applied to traditional data sources such as news articles and newsgroup documents. In this work, we introduce a simple approach based on sentence filtering in order to improve topical aspects learned from newsgroups-based content without modifying the basic mechanism of ABAE. We train a probabilistic classifier to distinguish between out-of-domain texts (outer dataset) and in-domain texts (target dataset). Then, during data preparation we filter out sentences that have a low probability of being in-domain and train the neural model on the remaining sentences. The positive effect of sentence filtering on topic coherence is demonstrated in comparison to aspect extraction models trained on unfiltered texts.
\end{abstract}

\end{frontmatter}

\section{Introduction}\label{intro}

Aspect extraction is an important task in sentiment analysis of, e.g., user reviews. The goal of aspect extraction is twofold: (1) to extract words/tokens describing features of the item the author shares their opinion about, (2) to attribute each extracted word/token to a group/cluster related to a certain feature. For example, given a sentence ``The stew was delicious'' from a restaurant review, one might extract  ``stew'' as an aspect word representing the ``food'' aspect. The words ``steak'', ``borscht'', ``fish'' etc. could also be attributed to the aspect ``food''.

Recent advances in neural attention-based architectures have made it into a method of choice for modern natural language processing. It is currently well established (see, e.g., \cite{abae,luo2019unsupervised}) that neural models are able to identify latent topical aspects in user-generated texts in an unsupervised way. The purpose of topic modeling is to cluster words (generally speaking, tokens) of the input text into coherent \textit{topics}, or \emph{aspects}; e.g., the words \textit{criminal} and \textit{federal} are part of the topic \textit{justice} for the domain of law journals, while \textit{oxidation} and \textit{reaction} are part of the topic \textit{chemistry} for the domain of research papers. In probabilistic topic models, the topics/aspects are usually defined as distributions over words/tokens; the topic distribution can then serve as a compressed description of the document for other models.

Unsupervised methods for aspect extraction and topic modeling are an active field of research, especially since they can be applicable to texts in any domain. In particular, one model that has recently proved to be successful is the unsupervised neural attention-based aspect extraction model (ABAE)~\cite{abae}. One of the most prominent advantages of attention-based models over traditional topic modes such as latent Dirichlet allocation (LDA)~\cite{BNJ03} is that the former encode word-occurrence statistics into word embeddings and apply an attention mechanism to remove irrelevant words, learning a set of aspect embeddings.

While recent studies on a set of user reviews have demonstrated that neural attention models can provide aspects of significantly higher quality than
the classical LDA model or its modifications developed over the last decade (see, e.g., \cite[Fig.~2]{abae}), we have found in 
recent research and 
practical experience (see Table~\ref{tab:examples} for an example) that these models have significant limitations on long texts such as, e.g., 
newsgroup posts as compared to user reviews on \emph{Amazon} or similar.

One possible explanation for this effect lies in the style differences between the two domains. Review writers expressing opinions and describing items of value (whether those are venues, goods, events, or anything else) usually stay focused on the topic and do not venture into general exposition. This means that the implicit assumption of a sentence-based model such as ABAE~\cite{abae} that every sentence relates to a single aspect usually makes sense.

On the contrary, newsgroup texts or longer reviews are ``too general'' compared to \emph{Amazon}-like reviews, i.e., too many sentences are ``general'' (see sentence~$3$ in Table~\ref{tab:sent-scores} for an example) and do not contain aspect words that ABAE or similar models are implicitly trained to encode and recover. 
 
The attention mechanism imposes restrictions on the model: ABAE learns a poor representation of texts at the broad, general level rather than in terms of latent topics discovered from the collection. For a prolonged example, Table~\ref{tab:examples} shows sample topical aspect words extracted for the \emph{sci.electronics} newsgroup from the \emph{20 Newsgroups} dataset. Each row in Table~\ref{tab:examples} contains eight most probable words for the corresponding aspect extracted by the ABAE model~\cite{abae}. In the left column, Table~\ref{tab:examples} shows examples of poor topical aspects learned by directly applying ABAE on all sentences of newsgroup documents and topic words that are much more coherent and readily interpretable.

\begin{table*}[t]
\small
\caption{Sample aspects extracted from \emph{sci.electronics} by ABAE~\cite{abae}. Left: aspects extracted from all sentences; right: aspects extracted from selected sentences}\label{tab:examples}
\begin{tabular}{l|l}
\toprule
\textbf{ABAE trained on all sentences in a post (less coherent)} & \textbf{ABAE trained on selected sentences (more coherent)} \\ \midrule

\textless num\textgreater~\textless pad\textgreater~ raffle anyone copy  & wiring green cable box gfci grounded case \\
time frequency chip source take much& voltage input supply output signal power circuit \\
\textless num\textgreater greggo \textless unk\textgreater  mc68882rc33 ~\textless pad\textgreater~ raffle &  edu university uk mail fax email internet\\ 
\textless unk\textgreater raffle \textless pad\textgreater greggo mc68882rc33 ~\textless num\textgreater~ ca input & dc digital wave drive per data decimal state  \\
dtmedin b30 catbyte ingr uunet com uucp look& com dtmedin b30 catbyte uunet ingr uucp al everywhere \\
mail edu university writes com email uk & radar detector number someone radio law shack \\
copy anyone know could would help get & ca mb bison baden inqmind de sys6626 mind bb bari\\
edu university uk henry toronto mail & best year around machine least seems band \\
input pin output data latch voltage & phone neoucom departmentedu oh usa computer uhura  \\
input output data voltage pin high & pin input latch output data voltage supply\\
ca mb bison baden inqmind de sys6626 bb & ground wire neutral conductor box outlet grounding  \\
connected outlet hot wire grounding neutral & uk mail university email com edu fax internet  \\
ground wire conductor neutral outlet connected & would anyone know copy get could want\\
mc68882rc33 \textless pad\textgreater greggo input raffle voltage & pin input neutral voltage connected wire current\\
phone neoucom edu department oh usa computer service & copy anyone know could would help  \\ \bottomrule
\end{tabular}
\end{table*}

How can we extract better aspects in longer and more general texts, e.g., in newsgroup posts, with standard ABAE? In this work we propose an intuitive solution to this problem based on sentence filtering. 

The idea is to train a simple binary probabilistic text classifier able to separate the texts of a particular (target) domain of interest from texts on other topics. For example, all news' sentences about \textit{sport} are labeled as in-domain texts for the target domain `sport', while texts about \textit{politics}, \textit{electronics} or \textit{weather} are considered as out-of-domain examples. This binary classifier allows to estimate the probability of each sentence to be an ``in-domain'' sentence in the target dataset. Sentences with scores lower than a certain threshold can then be treated as ``out-of-domain'' (general) ones and dropped from the training set for aspect extractors. Note that sentence classification here is not a goal in itself but serves as preprocessing for subsequent aspect extraction.

In this work we show that this simple technique allows to achieve better interpretability of the resulting aspects. For a clear example, see the right column of Table~\ref{tab:examples} that contains top words for aspects also extracted by ABAE but after the proposed preprocessing. Note that token sets in the two columns intersect often, but aspects in the left column are ``noisier'', less coherent, and harder to interpret. The aspects become better as the model is not trying to encode sentences that could be attributed to any other domain and is free to concentrate on ``relevant'' sentences.

The paper is structured as follows. Section \ref{sec:related} briefly surveys related work. In Section~\ref{sec:neural}, we begin with the model description, describing attention mechanisms and the existing ABAE model. In Section~\ref{sec:filtering}, we present an approach to sentence filtering using out-of-domain classification. The experimental setup and results on several datasets are presented in Section~\ref{sec:eval}. We conclude with a summary of our results and possible future research directions in Section~\ref{sec:concl}.

\section{Related work}\label{sec:related}

Topic modeling is a set of techniques intended to uncover the topical structure of a corpus of documents in an unsupervised manner; it has become the method of choice for a number of applications dealing with general text-level analysis. The most popular basic model is Latent Dirichlet Allocation (LDA)~\cite{BNJ03}, and over the last decade and a half it has given rise to numerous extensions and generalizations. Various topic models have been applied to many kinds of documents, including research abstracts, newspaper archives, \emph{Wikipedia} articles, user reviews, tweets, and other user-generated texts~\cite{BNJ03,GS04,CB10,WC06,WBH08,BM07,mehrotra2013improving,loukachevitch2018thesaurus}.

Studies that are the nearest to our present work in terms of novel approaches for input (pre)processing without modifying the generative process of the probabilistic models themselves include, e.g.,~\cite{mehrotra2013improving,krasnashchok2018improving,loukachevitch2018thesaurus}. In these studies, discovered topics were quantitatively evaluated in terms of topic coherence. Mehrotra et al. proposed a novel method of tweet pooling by hashtags in order to improve LDA topics~\cite{mehrotra2013improving}. Tweets were aggregated into ``macro-documents'', and the macro-documents were used as training data to construct better LDA models. First, all tweets were pooled by existing hashtags. Second, unlabeled tweets were assigned with hashtags if the similarity score between an unlabeled and labeled tweet exceeds a confidence threshold ($0.5$ in~\cite{mehrotra2013improving}). The similarity score was based on TF or TF-IDF vector space representations. The authors concluded that the novel scheme of hashtag-based pooling leads to drastically improved topic modeling as compared to unpooled tweets, author-wise, or time-wise pooled. Krasnashchok and Jouili employed a term-weighting approach for the LDA input in order to promote named entities~\cite{krasnashchok2018improving}. The authors artificially modified the frequencies of named entities in the \emph{20 Newsgroups} dataset without changing the weights of other terms. Experiments in the paper demonstrated that the proposed approach positively influences the overall topic quality. Loukachevitch et al. proposed a novel approach of computing word frequencies to use for LDA input based on thesaurus relations~\cite{loukachevitch2018thesaurus}. They hypothesised that if words from the same similarity set co-occur in the same document then their contribution into the document's topics is higher, therefore their frequencies should be increased. The results showed that document frequencies really do influence the coherence of topic models, and the proposed approach improves it.

In this work, we concentrate on the ABAE model~\cite{abae}. Since it was put forward in 2017, recent studies have utilized ABAE for various NLP tasks including rating prediction~\cite{aspera} and user profiling~\cite{mitcheltree2018using}. Unsupervised aspect extraction models such as ABAE~\cite{abae} are shown to yield interpretable and coherent aspects for the reviews of various goods (usually tested on the \emph{Amazon} reviews dataset), which are typically short and very focused on certain items of interest of the reviewer. Researchers from the \emph{Airbnb} team applied ABAE to a large corpus of accommodation reviews in order to generate review summaries and user profiles~\cite{mitcheltree2018using}. They evaluated ABAE across these two tasks. For the first task of extractive summarization, they used sentence-level aspects inferred by ABAE to select representative review sentences for a given accommodation and a given aspect. 
For the second task, the authors used sentence-level aspects to compute user profiles by grouping all reviews coming from a given user. 

Quantitative and qualitative analysis conducted in~\cite{mitcheltree2018using} showed that these user profiles are effective in reranking reviews and accommodations. Interestingly, the authors found that aspects inferred by the $k$-means baseline are relatively incoherent compared to ABAE. The $k$-means baseline works very well to identify frequent aspects, while ABAE is better for infrequent aspects.

Another recent model, Aspect-based Rating Prediction (\emph{AspeRa}), has been proposed in~\cite{aspera} for learning rating- and text-aware recommender systems based on neural attention-based aspect extraction produced by the ABAE model, metric learning, and autoencoder-enriched learning. The proposed model outperformed state of the art aspect-based recommender systems on several real-world datasets of user reviews. Moreover, aspects discovered by AspeRa as a side product of the rating prediction task proved to be readily interpretable and, when evaluated in terms of standard topic coherence metrics, showed quality similar to LDA.

\section{Neural architecture for aspect extraction}\label{sec:neural}

\subsection{Attention mechanisms}\label{sec:attn}

Attention mechanisms had initially appeared in computer vision, but were quickly adapted to recurrent architectures used for natural language processing. There, attention mechanisms were introduced to overcome a commonly known flaw of RNNs, the lack of long-term memory: without additional modifications, RNNs can quickly forget early timesteps~\cite{kirkpatrick2017overcoming}. Attention serves as a kind of recall mechanism, allowing the network to recall different parts of the input when necessary. The already classical approach to attention was defined in~\cite{bahdanau2014neural}. A more recent and advanced version of attention, known as the Transformer, was presented in~\cite{vaswani2017attention} and has  already served as a basis for many extensions; the general idea of \textit{self-attention} that we shall discuss further is extensively employed in that work.

\def\x{\boldsymbol{x}}
\def\w{\boldsymbol{w}}
\def\y{\boldsymbol{y}}
\def\h{\boldsymbol{h}}
\def\bc{\boldsymbol{c}}
\def\f{\boldsymbol{f}}
\def\b{\boldsymbol{b}}
\def\z{\boldsymbol{z}}
\def\bu{\boldsymbol{u}}
\def\bp{\boldsymbol{p}}
\def\br{\boldsymbol{r}}
\def\bv{\boldsymbol{\upsilon}}

The basic idea could be described as choosing the most ``interesting'' or ``relevant'' part of the input sequence to produce the current step of the output/the values in the next network layer. A soft alignment model produces \emph{attention weights} $a_i$ that control how much each input word influences the word currently being produced. The score $a_i$ indicates whether the network should be focusing on this specific word right now, and $\z_s$ is the text vector that summarizes all information from the words. Since attention is soft ($a_i$ are real numbers), the gradients are able to flow through the entire network, and the model can be trained end-to-end. Soft attention drastically improves translation (see \cite{bahdanau2014neural}) and other tasks, allowing recurrent architectures to operate with longer sentences than without it; it is now a standard approach.

More formally, the basic attention mechanism is defined as

\begin{align*}
    a_{i}&=\frac{\exp{(\w_i^\top \y_k)}}{\sum_{j=1}^n\exp{(\w_j^\top \y_k)}},\\
    \z_k&=\sum_{i=1}^n a_i \w_i,
\end{align*}
where 
$\y_k$ is the \textit{key vector} produced separately (we discuss it in the case of ABAE in the next section); intuitively, $\y_k$ represents the context, meaning that vectors which are closer to the current context one should have more weight; $\{ \w_i \}_{i=1}^n$ are the \emph{value vectors} from which one constructs $\z_k$, and $n$ is the number of words in the input. In case of ABAE and many other NLP models, the value vectors are sets/sequences of word embeddings corresponding to words from the input text.

\subsection{Neural attention-based aspect extraction model}\label{sec:abae}

\begin{figure}[t]\centering
\includegraphics[width=0.94\linewidth]{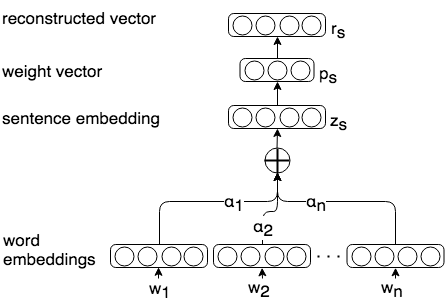}

\caption{Architecture of the the neural attention-based aspect extraction model (ABAE)~\cite{abae}.}\label{fig:abae}
\end{figure}

ABAE, the Neural Attention-Based Aspect Extraction Model~\cite{abae} is a neural architecture intended to capture the topical content of input texts. Similar to classical topic modeling~\cite{BNJ03}, the user chooses a finite number of topics (called \emph{aspects} in this context), and the goal of ABAE is to learn the aspects themselves and the extent to which each document corresponds to each of the aspects.

In essence, the ABAE model is an autoencoder; the primary component of the ABAE loss function is the reconstruction loss between the (weighted) sum of word embeddings used as the sentence representation and a linear combination of aspect embeddings. The sentence embedding is weighted by the so-called \emph{self-attention}, an attention mechanism where the values are embeddings of words in a sentence and the key $\y_s$ is the mean embedding of the same words.

Figure~\ref{fig:abae} illustrates the ABAE model in more detail. The first step for each sentence $s$ is to compute the sentence's embedding $\z_s \in \mathbb{R}^d$. In order to do this, 
for each word $w_i$ one retrieves a pre-trained word embedding $\w_i$, $\w_i \in \mathbb{R}^d$.

Then we compute attention weights $a_i$ as a multiplicative self-attention model:
$$
a_i = \frac{\exp{(\w_i^\top A \y_s)}}{\sum_{j=1}^n\exp{(\w_j^\top A \y_s)}},$$
where
$\y_s = \frac{1}{n}\sum_{i=1}^n \w_i$.
Here $A \in \mathbb{R}^{d \times d}$ is a matrix to be learned during end-to-end training. Importantly, the attention mechanism in ABAE is slightly different from the one described above in Section~\ref{sec:attn}; here, a simple dot product $\w_i^\top \y_s$ is replaced by a more complex bilinear transformation with a trained matrix $\w_i^\top A \y_s$. This modification does not change the dimension of the output vector and improves the model's expressive power. 

Once one has computed the attention weights, one computes the text representation $\z_s$ as a weighted sum of word embeddings:
  $$
    \z_s = \sum_{i=1}^n a_i \w_i.
  $$

The next step is to compute the aspect-based sentence representation $\br_s \in \mathbb{R}^d$ from an aspect embeddings matrix $T \in \mathbb{R}^{k \times d}$, where $k$ is the number of aspects:
  $$
    \br_s = T^{\top} \bp_s,\text{ where }
    \bp_s = \mathrm{softmax}(W\z_s + \b).
  $$
Here $\bp_s \in \mathbb{R}^k$ is the vector of probability weights over $k$ aspect embeddings, and $W \in \mathbb{R}^{k \times d}$, $\b\in\mathbb{R}^k$ are the parameters of a feed-forward layer.

Each of $k$ rows in matrix $T$ represents a ``topic embedding''. The original work by He~et~al.~\cite{abae} suggests to initialize it with centroids of pre-trained word vectors clusters, grouped with the $k$-means algorithm~\cite{steinhaus1956division,macqueen1967some,lloyd1982least}.

To train the model, ABAE defines the reconstruction error as the cosine distance between $\br_s$ and $\z_s$ with a contrastive max-margin objective function~\cite{weston2011wsabie}. In addition, an orthogonality penalty term is added to the objective, which tries to learn the aspect embedding matrix $T$ that would produce aspect embeddings that would be as diverse as possible. The entire architecture at a certain level of abstraction is presented in Fig.~\ref{fig:abae}.

\section{Approach}\label{sec:filtering}

As we have briefly outlined in the introduction, for longer texts such as newsgroup posts or articles we propose to select only certain sentences for training an unsupervised aspect extraction model. 

Let us consider the case when we have a target collection of newsgroups (or other texts longer than the average user review) of one certain domain ($ID$ for ``in-domain''). For our preprocessing approach we propose to do the following:
\begin{enumerate}[(1)]
    \item obtain a collection of out-of-domain texts $OOD$, split them into sentences;
    \item label the sentences from the target collection $ID$ as the ``in-domain'' class;
    \item label the sentences from the $OOD$ as the ``out-of-domain'' class;
    \item train a probabilistic binary classifier separating ``in-domain'' and ``out-of-domain'' classes;
    \item compute the ``probabilities'' (classifier scores) of each of the sentences from the target collection $ID$;
    \item choose a probability threshold $\theta$ and remove sentences that have a lower value of the probabilities computed above from the training set;
    \item train the unsupervised aspect extraction model (ABAE) on the filtered dataset $ID^f$.
\end{enumerate}

The procedure described above is very general. We could use any probabilistic classifier on steps (4)-(5), adopt any hyperparameter tuning scheme, and use different strategies for choosing the threshold for filtering the sentences. Note that we do not specify explicitly how exactly the out-of-domain data should be collected. As usual in modern natural language processing, we assume that such data can easily be collected on-demand and can include arbitrary texts. Therefore, although the classifiers are obviously trained in a supervised way, overall the proposed approach does not require any additional labeling and does not violate the unsupervised nature of aspect extraction.

In the next section, we describe the details of the exact approach used in our experiments and show our evaluation results.

\section{Experimental evaluation}\label{sec:eval}

\subsection{Evaluation metrics}\label{ssec:metrics}

In all experiments, we have evaluated the topics produced by ABAE and other topic models in our comparison in terms of topic coherence. The idea behind topic coherence is that a coherent topic will display words that tend to occur in the same documents. In other words, the most likely words in a coherent topic should have high mutual information. Document models with higher topic coherence are supposed to be the topic models with better interpretability.

We have employed standard topic coherence metrics:

\def\pmi{\mathrm{PMI}}
\def\npmi{\mathrm{NPMI}}
\def\cuci{C_{\mathrm{UCI}}}
\def\cpmi{C_{\mathrm{PMI}}}
\def\cnpmi{C_{\mathrm{NPMI}}}

\begin{enumerate}[(1)]
    \item $\cpmi$ (PMI-coherence)~\cite{newman2009external,Newman2010}: having taken top $N$ words from a topic/aspect, compute the average sum of PMIs for all $\frac{ (N-1)N }{ 2}$ pairs of words in the top $N$, where the probabilities in PMI are estimated as a smoothed frequency of co-occurrence in a sliding window:
    $$ \cpmi = \frac{2}{N(N-1)} \sum_{i=1}^{N-1} \sum_{j=i+1}^N \pmi(w_i, w_j),$$ 
    where
    $$\pmi(w_i, w_j) = \log{\frac{P(w_i, w_j) + \epsilon}{P(w_i)P(w_j)} };$$
    \item $\cnpmi$~\cite{bouma2009normalized}; this metric is similar to $\cpmi$ but employs the normalized PMI measure:
    $$ \npmi(w_i,w_j) = \left(\frac{\pmi(w_i, w_j)}{-\log{P(w_i, w_j) + \epsilon}}\right)^\gamma.$$
\end{enumerate}

In both metrics, we compute probabilities using co-occurrence frequencies within a sliding window of $10$ words. 

\subsection{Dataset}

To demonstrate the feasibility of our approach, in our experiments we have used the benchmark \emph{20~Newsgroups} dataset\footnote{\url{http://qwone.com/~jason/20Newsgroups/}}, which is essentially a collection of discussions on $20$~selected topics (newsgroups). We consider this dataset as a diverse collection of documents. Figure~\ref{fig:eval} shows the complete list of selected newsgroups.

Each newsgroup represents a certain domain. For each, we removed all meta-information describing the messages and all quotations of previous messages. We have split all the messages into sentences using the NLTK \cite{nltk} \texttt{sent\_tokenizer} and tokenized and normalized the terms in each sentence using the TweetTokenizer and WordNetLemmatizer~\cite{nltk}, respectively. 

For each newsgroup category, we have carried out the procedure described in Section~\ref{sec:filtering}. 

Every newsgroup's sentences were labeled as the ``in-domain'' class. All other newsgroups' sentences in the preprocessed \emph{20~Newsgroups} dataset were considered an out-of-domain collection (not related to the particular domain) and labeled as ``out-of-domain''. E.g. when preparing sentences for aspect extraction for $sci.electronics$, the texts of this newsgroup were treated as $ID$, and all other newsgroups $alt.atheism$, $misc.forsale$, etc. were concatenated and treated as $OOD$ set.

As stated in Section~\ref{sec:filtering}, we require a probabilistic classifier for sentence selection. Despite there is a vast variety of advanced text classification methods, we have decided to adopt a very straightforward approach to demonstrate the feasibility of the general procedure proposed in this study. Hence, we have decided to present each sentence as a bag-of-words representation and use logistic regression as the probabilistic classifier, adopting the \emph{scikit-learn} implementation~\cite{scikit-learn}. The classifier was trained until convergence with the maximum number of iterations equal to~$100$. We have used all of $ID$ and $OOD$ for each newsgroup as train set. Since out-of-domain classification itself is not the main point of this work, we did not evaluate the classifiers predictions on any test sets. We note that we acknowledge that the classifiers' quality may influence the overall results and leave this analysis in for future analysis. 

The evaluation results of the binary classifier on the training sets are presented in Table~\ref{tab:gof}. The results show that the model based on logistic regression and bag-of-words representations obtained 94\%-98\% accuracy. We also present samples of sentences with the obtained scores from the \emph{sci.electronics} newsgroup in Table~\ref{tab:sent-scores}.

\begin{table*}[h]
\small
\caption{\label{font-table} The evaluation results of the binary classifier on the training sets for each newsgroup}\label{tab:gof}
\begin{tabular}{|l|lll|}
\hline
\textbf{Newsgroup ($ID$)} & \textbf{Precision} & \textbf{Recall} &  \textbf{Accuracy} \\
\hline
sci.electronics &   0.89    &  0.19 &     0.97 \\
soc.religion.christian  &    0.82   &   0.36  &    0.95\\
rec.sport.baseball  &    0.92  &    0.36  &    0.97\\
comp.sys.ibm.pc.hardware   &   0.79   &   0.22    &  0.97\\
misc.forsale  &    0.87   &   0.29  &    0.98\\
alt.atheism    &  0.76  &    0.18  &    0.95\\
sci.med  &    0.94   &   0.35   &  0.96\\
talk.politics.misc  &    0.86   &   0.22   &   0.94\\
\hline
\end{tabular}
\end{table*}

\begin{table*}[t]
    \centering
    \caption{Out-of-domain classifier's scores for sentences from the \emph{sci.electronics} newsgroup}
    \label{tab:sent-scores}
    \begin{tabular}{|l|l|}
        \hline
        \textbf{Score} & \textbf{Sentence after preprocessing with NLTK} \\ \hline
        0.844 & paul simundza writes probably tell dc blocking capacitor series one chip single ended audio amp speaker terminal \\ 
        0.836 & open look power amp ic \\
        0.047 & fairly obvious \\ 
        0.466 & replace one connected dead output \\ 
        0.668 & well one thing poke around terminal power amp chip \\   \hline
    \end{tabular}
\end{table*}

After applying the classifier, we have generated new datasets by filtering each of the chosen newsgroups by every score threshold from the set $\{0.0$, $0.1, 0.2, ..., 0.9\}$, where $0.0$ means no filtering. As the threshold increases, the datasets are reduced in size; for example, for threshold $0.5$ the \emph{Christianity} ($soc.religion.christian$) dataset size is reduced by 55\%.

\subsection{Experimental setup}

Following ABAE~\cite{abae}, we set the ortho-regularization coefficient for the aspect matrix equal to $0.1$. Since this model utilizes an aspect embedding matrix to approximate aspect words in the vocabulary, initialization of aspect embeddings is crucial. We adopted the approach described in the original work~\cite{abae}, initializing based on \emph{k-means} clustering~\cite{steinhaus1956division,macqueen1967some,lloyd1982least}. In this method, all word vectors (i.e., \emph{word2vec}~\cite{MCCD13}) for the words occurring in input texts are clustered with $k$-means, and then
rows of the aspect embedding matrix are initialized with centroids of the resulting clusters. We have used $15$ aspects (topics) and $20$ negative samples for learning phase following \cite{abae}. We trained the model for $10$ epochs with a batch size of $256$ on one GPU. 

ABAE is initialized with the \emph{word2vec} (\emph{SGNS}) vectors, trained on the corresponding domain (newsgroup) for every newsgroup with the following settings: the dimension is $200$, the window size equals $10$, the number of negative samples equals $5$, and only words with the minimal count of $2$ are taken into account. We used the \emph{gensim} library~\cite{rehurek2010gensim} to train the \emph{SGNS} models. 
We adopted the \emph{OnlineLDA} model~\cite{hoffman2010online} trained with the \emph{gensim} library~\cite{rehurek2010gensim} with default parameters, using the same vocabulary and the same number of aspects as in ABAE.

\subsection{Results}

We have trained ABAE with sentences as input (as in the original paper \cite{abae}), using the filtered datasets generated as shown above. 

The models we used for comparison as baselines are:
\begin{enumerate}[(1)]
    \item ABAE trained on full texts of posts in the newsgroups;
    \item OnlineLDA trained on full texts of posts;
    \item OnlineLDA trained on sentences.
\end{enumerate}

For every dataset, we have trained an aspect extraction model and computed two coherence metrics defined above using the software accompanying the paper~\cite{lau2014machine}.

\begin{figure*}[!t]\centering
\scalebox{.90}{
\setlength{\tabcolsep}{6pt}
\begin{tabular}{p{.5\linewidth}p{.5\linewidth}}

\includegraphics{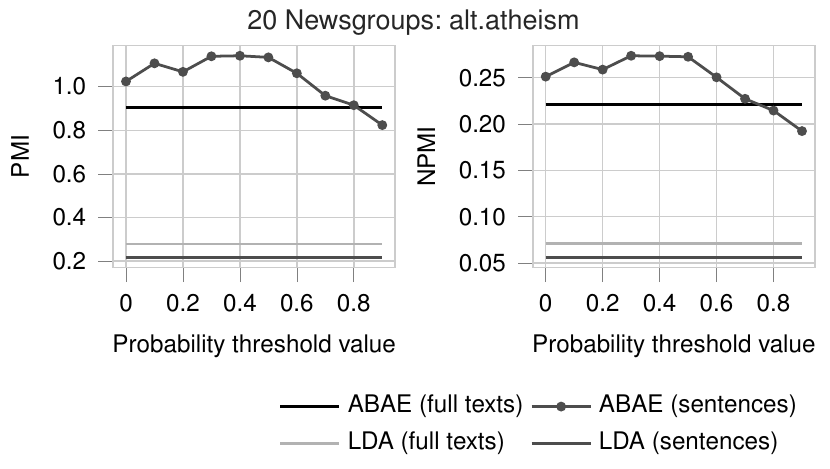}

&
\includegraphics{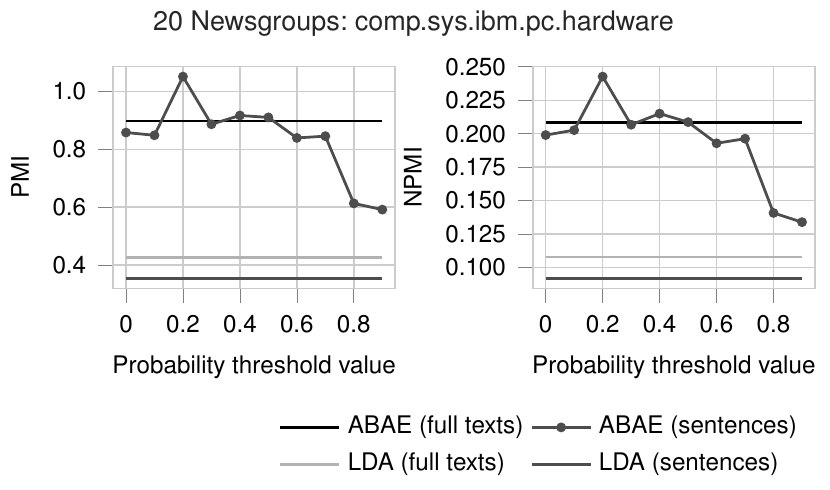}
\\[5mm]

\includegraphics{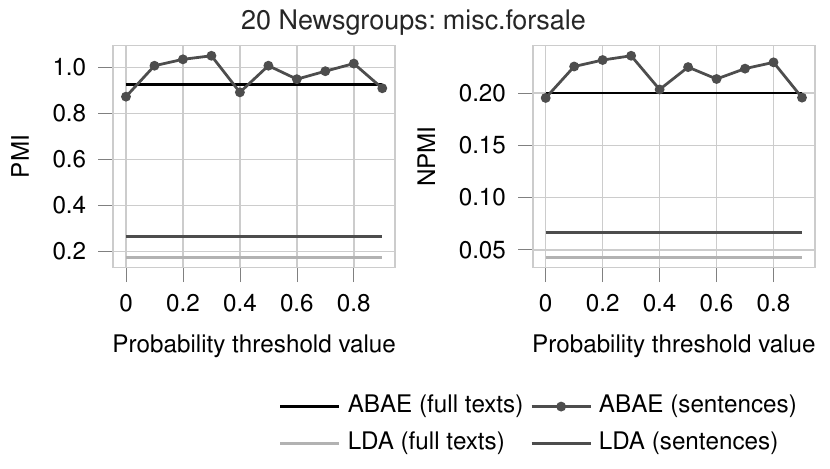}
&
\includegraphics{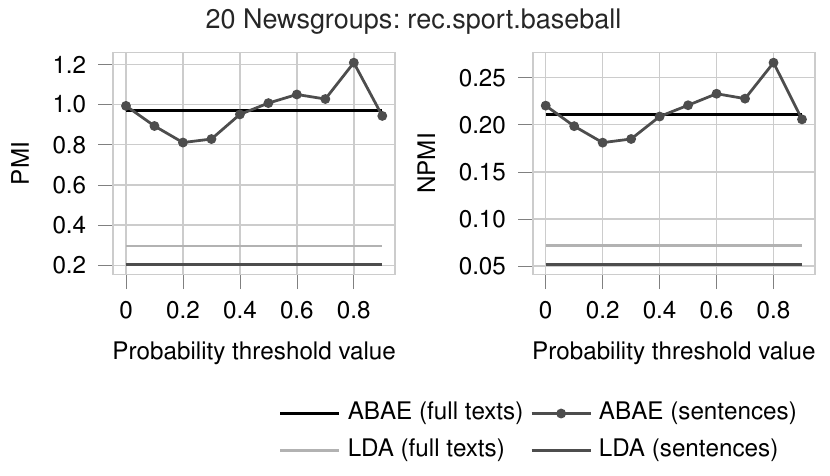}
\\[5mm]

\includegraphics{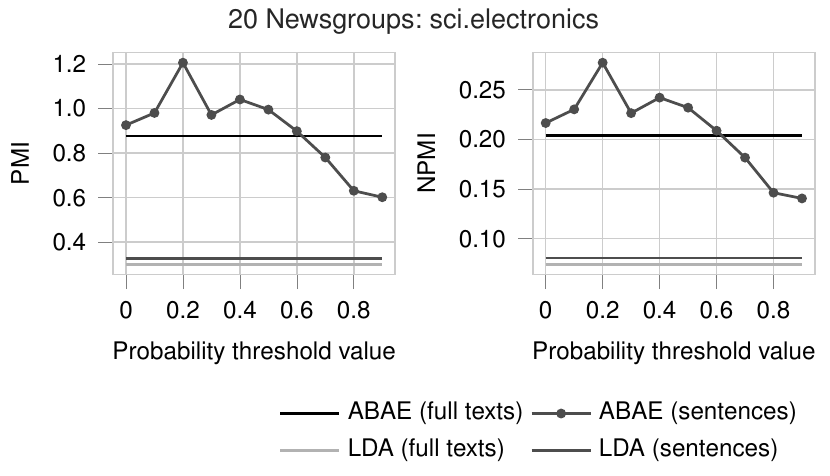}
&
\includegraphics{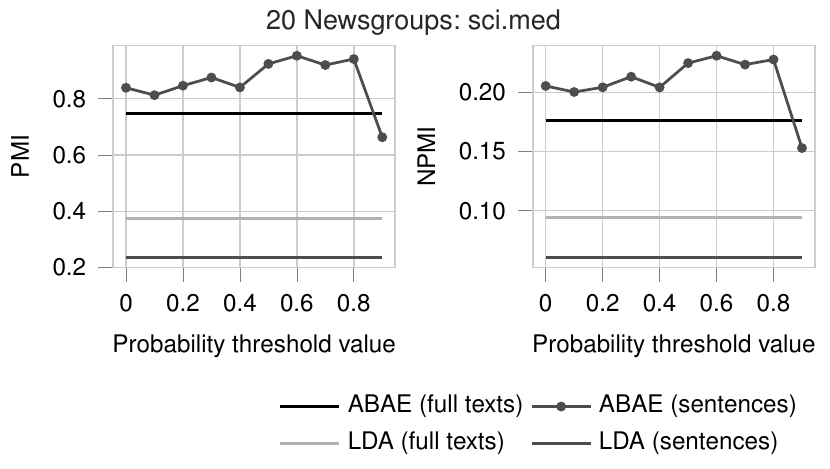}
\\[5mm]

\includegraphics{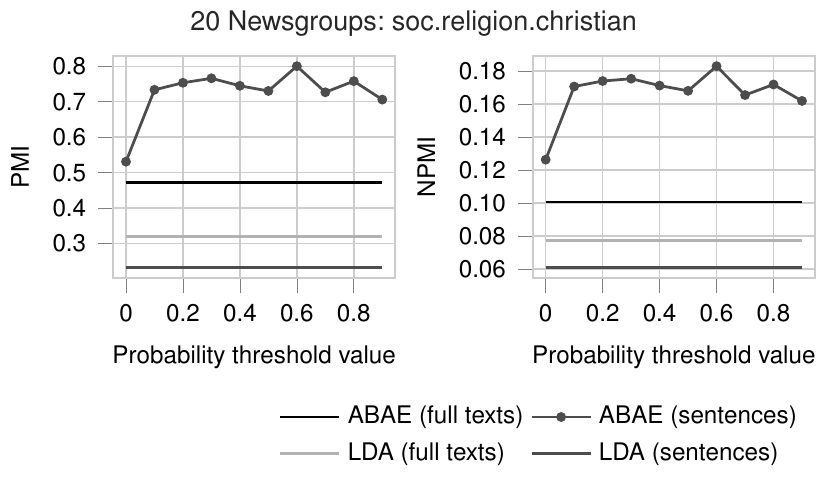}
&
\includegraphics{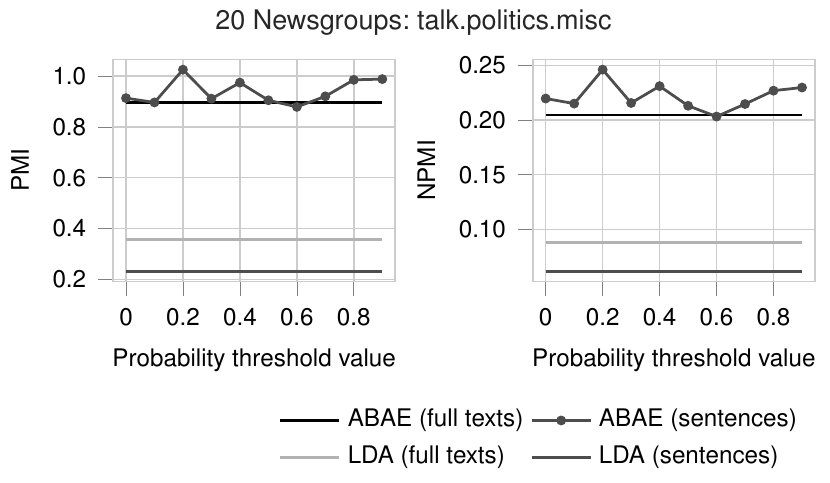}
\\[5mm]

\end{tabular}
}
\caption{Evaluation of aspects extracted from the \emph{20 Newsgroups} collection by different topic models.}\label{fig:eval}
\end{figure*}

Figure~\ref{fig:eval} contains the results across all datasets in the comparison. It clearly shows that in most cases it is possible to choose a filtering strategy to increase the topic coherence provided by the model. 

Several interesting observations can be made based on these figures. First, the optimal threshold varies for different domains, yet for the most domains the threshold $0.2$ increases coherence for extracted topics. Although there are exceptions (e.g. for the \emph{Baseball} domain the $0.2$ threshold does nothing), this fact needs further investigation. Generally, we can conclude that even this simplistic filtering technique improves the quality of an ABAE model with a reduction of data samples and therefore significantly reduced training time.

Second, the LDA baselines in comparison to each other show that full-text training data results in higher quality, which could be interpreted as proof that longer texts are more appropriate for the LDA model. Indeed, the LDA model generally was designed to work on texts longer than a typical sentence.

Third, interestingly, the ABAE model on full texts consistently shows better results than LDA baselines, despite the fact that it was designed to work on short texts (one or two sentences). Finally, one can clearly see that in all conducted experiments there is no significant difference in the results between PMI and NPMI coherence measures. 

\begin{figure}[!t]
    \centering
    \includegraphics[width=\linewidth]{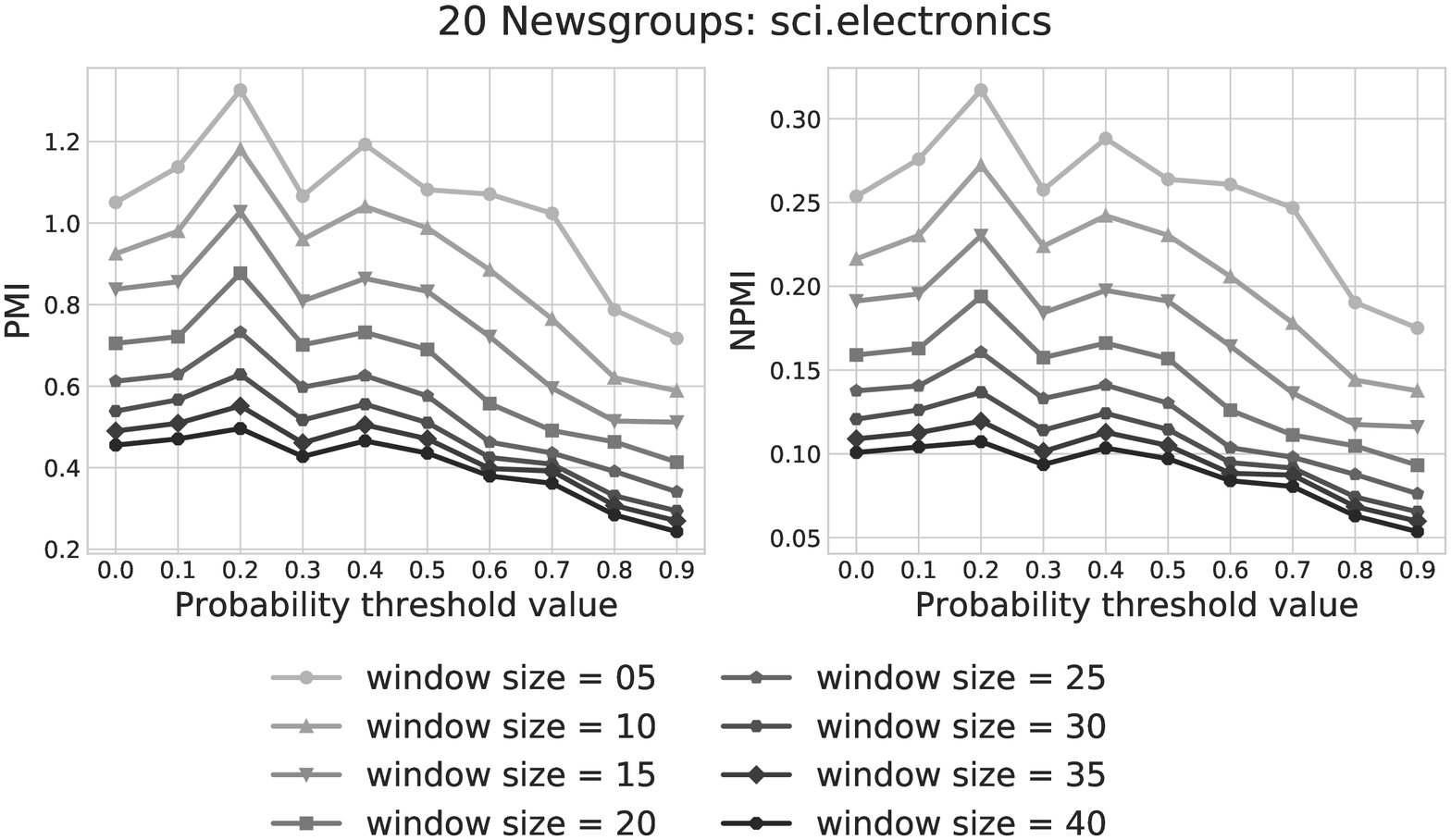}
    \caption{Topic coherence as a function of the probability threshold value for the \emph{sci.electronics} newsgroup for different window sizes.}\label{fig:window}
\end{figure}

We have also experimented with other window sizes but found that the general form of the PMI and NPMI curves remains the same for all reasonable window sizes; see Figure~\ref{fig:window} for an illustration.

\section{Conclusion}\label{sec:concl}

In this work, we have presented a simple yet effective method of filtering out-of-domain sentences in order to improve the quality of ABAE-based models in newsgroup posts in terms of topic coherence. The presented results on the \emph{20~Newsgroups} dataset demonstrate that the proposed filtering method indeed improves the overall topic quality: ABAE trained on in-domain sentences discovers better topics than both
\begin{enumerate}[(1)]
    \item LDA trained on either full texts or sentences and
    \item ABAE trained on both in-domain and out-of-domain sentences.
\end{enumerate}

We see several potential directions for future work. First, there are more sophisticated topic models than the basic LDA, which could be even more sensitive to in- and out-of-domain data. We posit that the proposed technique can help some of them even more. 

Second, another potential research direction could be to use more complex techniques for this in/out-of-domain classification, e.g., the method described in~\cite{ryu2018out}. In general, we feel that the proposed filtering approach is a universal technique that can bring improvements across different topic models and neural architectures.

Finally, although we consider our claim fully supported by the evidence provided in this work, to make the proposed technique practical one also has to devise a reliable method of choosing the threshold. The threshold clearly depends on both the dataset and out-of-domain classification models. As the models are yet to be compared (see above), the technique for choosing the threshold is left for further study as well.

\section*{Acknowledgments}
Work on problem definition and model development was carried out at the Samsung-PDMI Joint AI Center at PDMI RAS and supported by Samsung Research. We also thank the anonymous reviewers whose comments have allowed us to improve the paper.


\bibliographystyle{ios1}

\begin{thebibliography}{31}
\ifx \bisbn   \undefined \def \bisbn  #1{ISBN #1}\fi
\ifx \binits  \undefined \def \binits#1{#1} \fi
\ifx \bauthor  \undefined \def \bauthor#1{#1} \fi
\ifx \bjtitle  \undefined \def \bjtitle#1{\textit{#1}}\fi
\ifx \batitle  \undefined \def \batitle#1{#1} \fi
\ifx \bctitle  \undefined \def \bctitle#1{#1} \fi
\ifx \bvolume  \undefined \def \bvolume#1{\textbf{#1}}\fi
\ifx \byear  \undefined \def \byear#1{#1} \fi
\ifx \bissue  \undefined \def \bissue#1{#1} \fi
\ifx \bfpage  \undefined \def \bfpage#1{#1} \fi
\ifx \blpage  \undefined \def \blpage #1{#1} \fi
\ifx \burl  \undefined \def \burl#1{#1} \fi
\ifx \doiurl  \undefined \def \doiurl#1{#1} \fi
\ifx \betal  \undefined \def \betal{et al.} \fi
\ifx \binstitute  \undefined \def \binstitute#1{#1} \fi
\ifx \beditor  \undefined \def \beditor#1{#1} \fi
\ifx \bpublisher  \undefined \def \bpublisher#1{#1} \fi
\ifx \bbtitle  \undefined \def \bbtitle#1{\textit{#1}} \fi
\ifx \bedition  \undefined \def \bedition#1{#1} \fi
\ifx \bseriesno  \undefined \def \bseriesno#1{#1} \fi
\ifx \blocation  \undefined \def \blocation#1{#1} \fi
\ifx \bsertitle  \undefined \def \bsertitle#1{#1} \fi
\ifx \bsnm \undefined \def \bsnm#1{#1} \fi
\ifx \bsuffix \undefined \def \bsuffix#1{#1} \fi
\ifx \bparticle \undefined \def \bparticle#1{#1} \fi
\ifx \barticle \undefined \def \barticle#1{#1} \fi
\ifx \botherref \undefined \def \botherref #1{#1} \fi
\ifx \url \undefined \def \url#1{#1} \fi
\ifx \bchapter \undefined \def \bchapter#1{#1} \fi
\ifx \bbook \undefined \def \bbook#1{#1} \fi
\ifx \bcomment \undefined \def \bcomment#1{#1} \fi
\ifx \oauthor \undefined \def \oauthor#1{#1} \fi
\ifx \citeauthoryear \undefined \def \citeauthoryear#1{#1} \fi
\ifx \texttildelow  \undefined \def \texttildelow{\symbol{126}} \fi
\def \endbibitem {}
\ifx \bconflocation  \undefined \def \bconflocation#1{#1} \fi

\bibitem[\protect\citeauthoryear{Bahdanau et~al.}{2014}]{bahdanau2014neural}
\begin{botherref}
\oauthor{\binits{D.}~\bsnm{Bahdanau}},
\oauthor{\binits{K.}~\bsnm{Cho}},
\oauthor{\binits{Y.}~\bsnm{Bengio}} and
\oauthor{\binits{R.}~\bsnm{Aharoni}},
Neural Machine Translation by Jointly Learning to Align and Translate,
\textit{Proceedings of International Conference of Learning Representation}
(2014).
\end{botherref}
\endbibitem

\bibitem[\protect\citeauthoryear{Bird et~al.}{2009}]{nltk}
\begin{bbook}
\bauthor{\binits{S.}~\bsnm{Bird}},
\bauthor{\binits{E.}~\bsnm{Klein}} and
\bauthor{\binits{E.}~\bsnm{Loper}},
\bbtitle{Natural language processing with Python: analyzing text with the
  natural language toolkit},
\bpublisher{" O'Reilly Media, Inc."},
\byear{2009}.
\end{bbook}
\endbibitem

\bibitem[\protect\citeauthoryear{Blei and McAuliffe}{2007}]{BM07}
\begin{botherref}
\oauthor{\binits{D.M.}~\bsnm{Blei}} and
\oauthor{\binits{J.D.}~\bsnm{McAuliffe}},
Supervised Topic Models,
\textit{Advances in Neural Information Processing Systems}
\textbf{22}
(2007).
\end{botherref}
\endbibitem

\bibitem[\protect\citeauthoryear{Blei et~al.}{2003}]{BNJ03}
\begin{barticle}
\bauthor{\binits{D.M.}~\bsnm{Blei}},
\bauthor{\binits{A.Y.}~\bsnm{Ng}} and
\bauthor{\binits{M.I.}~\bsnm{Jordan}},
\batitle{Latent {D}irichlet allocation},
\bjtitle{Journal of Machine Learning Research}
\bvolume{3}(\bissue{4--5})
(\byear{2003}),
\bfpage{993}--\blpage{1022}.
\end{barticle}
\endbibitem

\bibitem[\protect\citeauthoryear{Bouma}{2009}]{bouma2009normalized}
\begin{botherref}
\oauthor{\binits{G.}~\bsnm{Bouma}},
Normalized (pointwise) mutual information in collocation extraction,
\textit{Proceedings of GSCL}
(2009),
31--40.
\end{botherref}
\endbibitem

\bibitem[\protect\citeauthoryear{Chang and Blei}{2010}]{CB10}
\begin{barticle}
\bauthor{\binits{J.}~\bsnm{Chang}} and
\bauthor{\binits{D.M.}~\bsnm{Blei}},
\batitle{Hierarchical Relational Models for Document Networks},
\bjtitle{Annals of Applied Statistics}
\bvolume{4}(\bissue{1})
(\byear{2010}),
\bfpage{124}--\blpage{150}.
\end{barticle}
\endbibitem

\bibitem[\protect\citeauthoryear{Griffiths and Steyvers}{2004}]{GS04}
\begin{barticle}
\bauthor{\binits{T.}~\bsnm{Griffiths}} and
\bauthor{\binits{M.}~\bsnm{Steyvers}},
\batitle{Finding Scientific Topics},
\bjtitle{Proceedings of the National Academy of Sciences}
\bvolume{101 (Suppl. 1)}
(\byear{2004}),
\bfpage{5228}--\blpage{5335}.
\end{barticle}
\endbibitem

\bibitem[\protect\citeauthoryear{He et~al.}{2017}]{abae}
\begin{bchapter}
\bauthor{\binits{R.}~\bsnm{He}},
\bauthor{\binits{W.S.}~\bsnm{Lee}},
\bauthor{\binits{H.T.}~\bsnm{Ng}} and
\bauthor{\binits{D.}~\bsnm{Dahlmeier}},
\bctitle{An unsupervised neural attention model for aspect extraction},
in: \bbtitle{Proceedings of the 55th Annual Meeting of the Association for
  Computational Linguistics (Volume 1: Long Papers)},
\byear{2017},
pp.~\bfpage{388}--\blpage{397}.
\end{bchapter}
\endbibitem

\bibitem[\protect\citeauthoryear{Hoffman et~al.}{2010}]{hoffman2010online}
\begin{bchapter}
\bauthor{\binits{M.}~\bsnm{Hoffman}},
\bauthor{\binits{F.R.}~\bsnm{Bach}} and
\bauthor{\binits{D.M.}~\bsnm{Blei}},
\bctitle{Online learning for latent dirichlet allocation},
in: \bbtitle{advances in neural information processing systems},
\byear{2010},
pp.~\bfpage{856}--\blpage{864}.
\end{bchapter}
\endbibitem

\bibitem[\protect\citeauthoryear{Kirkpatrick
  et~al.}{2017}]{kirkpatrick2017overcoming}
\begin{barticle}
\bauthor{\binits{J.}~\bsnm{Kirkpatrick}},
\bauthor{\binits{R.}~\bsnm{Pascanu}},
\bauthor{\binits{N.}~\bsnm{Rabinowitz}},
\bauthor{\binits{J.}~\bsnm{Veness}},
\bauthor{\binits{G.}~\bsnm{Desjardins}},
\bauthor{\binits{A.A.}~\bsnm{Rusu}},
\bauthor{\binits{K.}~\bsnm{Milan}},
\bauthor{\binits{J.}~\bsnm{Quan}},
\bauthor{\binits{T.}~\bsnm{Ramalho}},
\bauthor{\binits{A.}~\bsnm{Grabska-Barwinska}} \betal,
\batitle{Overcoming catastrophic forgetting in neural networks},
\bjtitle{Proceedings of the national academy of sciences}
\bvolume{114}(\bissue{13})
(\byear{2017}),
\bfpage{3521}--\blpage{3526}.
\end{barticle}
\endbibitem

\bibitem[\protect\citeauthoryear{Krasnashchok and
  Jouili}{2018}]{krasnashchok2018improving}
\begin{bchapter}
\bauthor{\binits{K.}~\bsnm{Krasnashchok}} and
\bauthor{\binits{S.}~\bsnm{Jouili}},
\bctitle{Improving Topic Quality by Promoting Named Entities in Topic
  Modeling},
in: \bbtitle{Proceedings of the 56th Annual Meeting of the Association for
  Computational Linguistics (Volume 2: Short Papers)},
\byear{2018},
pp.~\bfpage{247}--\blpage{253}.
\end{bchapter}
\endbibitem

\bibitem[\protect\citeauthoryear{Lau et~al.}{2014}]{lau2014machine}
\begin{bchapter}
\bauthor{\binits{J.H.}~\bsnm{Lau}},
\bauthor{\binits{D.}~\bsnm{Newman}} and
\bauthor{\binits{T.}~\bsnm{Baldwin}},
\bctitle{Machine reading tea leaves: Automatically evaluating topic coherence
  and topic model quality},
in: \bbtitle{Proceedings of the 14th Conference of the European Chapter of the
  Association for Computational Linguistics},
\byear{2014},
pp.~\bfpage{530}--\blpage{539}.
\end{bchapter}
\endbibitem

\bibitem[\protect\citeauthoryear{Lloyd}{1982}]{lloyd1982least}
\begin{barticle}
\bauthor{\binits{S.}~\bsnm{Lloyd}},
\batitle{Least squares quantization in PCM},
\bjtitle{IEEE transactions on information theory}
\bvolume{28}(\bissue{2})
(\byear{1982}),
\bfpage{129}--\blpage{137}.
\end{barticle}
\endbibitem

\bibitem[\protect\citeauthoryear{Loukachevitch
  et~al.}{2018}]{loukachevitch2018thesaurus}
\begin{bchapter}
\bauthor{\binits{N.}~\bsnm{Loukachevitch}},
\bauthor{\binits{K.}~\bsnm{Ivanov}} and
\bauthor{\binits{B.}~\bsnm{Dobrov}},
\bctitle{Thesaurus-Based Topic Models and Their Evaluation},
in: \bbtitle{Proceedings of the 8th International Conference on Web
  Intelligence, Mining and Semantics},
\binstitute{ACM},
\byear{2018},
p.~\bfpage{11}.
\end{bchapter}
\endbibitem

\bibitem[\protect\citeauthoryear{Luo et~al.}{2019}]{luo2019unsupervised}
\begin{bchapter}
\bauthor{\binits{L.}~\bsnm{Luo}},
\bauthor{\binits{X.}~\bsnm{Ao}},
\bauthor{\binits{Y.}~\bsnm{Song}},
\bauthor{\binits{J.}~\bsnm{Li}},
\bauthor{\binits{X.}~\bsnm{Yang}},
\bauthor{\binits{Q.}~\bsnm{He}} and
\bauthor{\binits{D.}~\bsnm{Yu}},
\bctitle{Unsupervised neural aspect extraction with sememes},
in: \bbtitle{Proc. 28th Int. Joint Conf. Artif. Intell},
\byear{2019},
pp.~\bfpage{5123}--\blpage{5129}.
\end{bchapter}
\endbibitem

\bibitem[\protect\citeauthoryear{MacQueen et~al.}{1967}]{macqueen1967some}
\begin{bchapter}
\bauthor{\binits{J.}~\bsnm{MacQueen}} \betal,
\bctitle{Some methods for classification and analysis of multivariate
  observations},
in: \bbtitle{Proceedings of the fifth Berkeley symposium on mathematical
  statistics and probability},
Vol.~\bseriesno{1},
\binstitute{Oakland, CA, USA},
\byear{1967},
pp.~\bfpage{281}--\blpage{297}.
\end{bchapter}
\endbibitem

\bibitem[\protect\citeauthoryear{Mehrotra et~al.}{2013}]{mehrotra2013improving}
\begin{bchapter}
\bauthor{\binits{R.}~\bsnm{Mehrotra}},
\bauthor{\binits{S.}~\bsnm{Sanner}},
\bauthor{\binits{W.}~\bsnm{Buntine}} and
\bauthor{\binits{L.}~\bsnm{Xie}},
\bctitle{Improving lda topic models for microblogs via tweet pooling and
  automatic labeling},
in: \bbtitle{Proceedings of the 36th international ACM SIGIR conference on
  Research and development in information retrieval},
\binstitute{ACM},
\byear{2013},
pp.~\bfpage{889}--\blpage{892}.
\end{bchapter}
\endbibitem

\bibitem[\protect\citeauthoryear{Mikolov et~al.}{2013}]{MCCD13}
\begin{botherref}
\oauthor{\binits{T.}~\bsnm{Mikolov}},
\oauthor{\binits{K.}~\bsnm{Chen}},
\oauthor{\binits{G.}~\bsnm{Corrado}} and
\oauthor{\binits{J.}~\bsnm{Dean}},
Efficient Estimation of Word Representations in Vector Space,
\textit{CoRR}
\textbf{abs/1301.3781}
(2013).
\url{http://arxiv.org/abs/1301.3781}.
\end{botherref}
\endbibitem

\bibitem[\protect\citeauthoryear{Mitcheltree
  et~al.}{2018}]{mitcheltree2018using}
\begin{bchapter}
\bauthor{\binits{C.}~\bsnm{Mitcheltree}},
\bauthor{\binits{V.}~\bsnm{Wharton}} and
\bauthor{\binits{A.}~\bsnm{Saluja}},
\bctitle{Using Aspect Extraction Approaches to Generate Review Summaries and
  User Profiles},
in: \bbtitle{Proceedings of the 2018 Conference of the North American Chapter
  of the Association for Computational Linguistics: Human Language
  Technologies, Volume 3 (Industry Papers)},
\byear{2018},
pp.~\bfpage{68}--\blpage{75}.
\end{bchapter}
\endbibitem

\bibitem[\protect\citeauthoryear{Newman et~al.}{2009}]{newman2009external}
\begin{bchapter}
\bauthor{\binits{D.}~\bsnm{Newman}},
\bauthor{\binits{S.}~\bsnm{Karimi}} and
\bauthor{\binits{L.}~\bsnm{Cavedon}},
\bctitle{External evaluation of topic models},
in: \bbtitle{in Australasian Doc. Comp. Symp., 2009},
\binstitute{Citeseer},
\byear{2009}.
\end{bchapter}
\endbibitem

\bibitem[\protect\citeauthoryear{Newman et~al.}{2010}]{Newman2010}
\begin{bchapter}
\bauthor{\binits{D.}~\bsnm{Newman}},
\bauthor{\binits{J.H.}~\bsnm{Lau}},
\bauthor{\binits{K.}~\bsnm{Grieser}} and
\bauthor{\binits{T.}~\bsnm{Baldwin}},
\bctitle{Automatic Evaluation of Topic Coherence},
in: \bbtitle{Human Language Technologies: The 2010 Annual Conference of the
  North American Chapter of the Association for Computational Linguistics},
\bsertitle{HLT '10},
\bpublisher{Association for Computational Linguistics},
\blocation{Stroudsburg, PA, USA},
\byear{2010},
pp.~\bfpage{100}--\blpage{108}.
ISBN \bisbn{1-932432-65-5}.
\url{http://dl.acm.org/citation.cfm?id=1857999.1858011}.
\end{bchapter}
\endbibitem

\bibitem[\protect\citeauthoryear{Nikolenko et~al.}{2019}]{aspera}
\begin{bchapter}
\bauthor{\binits{S.I.}~\bsnm{Nikolenko}},
\bauthor{\binits{E.}~\bsnm{Tutubalina}},
\bauthor{\binits{V.}~\bsnm{Malykh}},
\bauthor{\binits{I.}~\bsnm{Shenbin}} and
\bauthor{\binits{A.}~\bsnm{Alekseev}},
\bctitle{AspeRa: Aspect-Based Rating Prediction Model},
in: \bbtitle{Advances in Information Retrieval},
\beditor{\binits{L.}~\bsnm{Azzopardi}},
\beditor{\binits{B.}~\bsnm{Stein}},
\beditor{\binits{N.}~\bsnm{Fuhr}},
\beditor{\binits{P.}~\bsnm{Mayr}},
\beditor{\binits{C.}~\bsnm{Hauff}} and
\beditor{\binits{D.}~\bsnm{Hiemstra}}, eds,
\bpublisher{Springer International Publishing},
\blocation{Cham},
\byear{2019},
pp.~\bfpage{163}--\blpage{171}.
ISBN \bisbn{978-3-030-15719-7}.
\end{bchapter}
\endbibitem

\bibitem[\protect\citeauthoryear{Pedregosa et~al.}{2011}]{scikit-learn}
\begin{barticle}
\bauthor{\binits{F.}~\bsnm{Pedregosa}},
\bauthor{\binits{G.}~\bsnm{Varoquaux}},
\bauthor{\binits{A.}~\bsnm{Gramfort}},
\bauthor{\binits{V.}~\bsnm{Michel}},
\bauthor{\binits{B.}~\bsnm{Thirion}},
\bauthor{\binits{O.}~\bsnm{Grisel}},
\bauthor{\binits{M.}~\bsnm{Blondel}},
\bauthor{\binits{P.}~\bsnm{Prettenhofer}},
\bauthor{\binits{R.}~\bsnm{Weiss}},
\bauthor{\binits{V.}~\bsnm{Dubourg}},
\bauthor{\binits{J.}~\bsnm{Vanderplas}},
\bauthor{\binits{A.}~\bsnm{Passos}},
\bauthor{\binits{D.}~\bsnm{Cournapeau}},
\bauthor{\binits{M.}~\bsnm{Brucher}},
\bauthor{\binits{M.}~\bsnm{Perrot}} and
\bauthor{\binits{E.}~\bsnm{Duchesnay}},
\batitle{Scikit-learn: Machine Learning in {P}ython},
\bjtitle{Journal of Machine Learning Research}
\bvolume{12}
(\byear{2011}),
\bfpage{2825}--\blpage{2830}.
\end{barticle}
\endbibitem

\bibitem[\protect\citeauthoryear{Rehurek and Sojka}{2010}]{rehurek2010gensim}
\begin{bchapter}
\bauthor{\binits{R.}~\bsnm{Rehurek}} and
\bauthor{\binits{P.}~\bsnm{Sojka}},
\bctitle{{Software Framework for Topic Modelling with Large Corpora}},
in: \bbtitle{{Proceedings of the LREC 2010 Workshop on New Challenges for NLP
  Frameworks}},
\bpublisher{ELRA},
\blocation{Valletta, Malta},
\byear{2010},
pp.~\bfpage{45}--\blpage{50},
\bcomment{\url{http://is.muni.cz/publication/884893/en}}.
\end{bchapter}
\endbibitem

\bibitem[\protect\citeauthoryear{Ryu et~al.}{2018}]{ryu2018out}
\begin{bchapter}
\bauthor{\binits{S.}~\bsnm{Ryu}},
\bauthor{\binits{S.}~\bsnm{Koo}},
\bauthor{\binits{H.}~\bsnm{Yu}} and
\bauthor{\binits{G.G.}~\bsnm{Lee}},
\bctitle{Out-of-domain Detection based on Generative Adversarial Network},
in: \bbtitle{Proceedings of the 2018 Conference on Empirical Methods in Natural
  Language Processing},
\byear{2018},
pp.~\bfpage{714}--\blpage{718}.
\end{bchapter}
\endbibitem

\bibitem[\protect\citeauthoryear{Steinhaus}{1956}]{steinhaus1956division}
\begin{barticle}
\bauthor{\binits{H.}~\bsnm{Steinhaus}},
\batitle{Sur la division des corp materiels en parties},
\bjtitle{Bull. Acad. Polon. Sci}
\bvolume{1}(\bissue{804})
(\byear{1956}),
\bfpage{801}.
\end{barticle}
\endbibitem

\bibitem[\protect\citeauthoryear{Vaswani et~al.}{2017}]{vaswani2017attention}
\begin{bchapter}
\bauthor{\binits{A.}~\bsnm{Vaswani}},
\bauthor{\binits{N.}~\bsnm{Shazeer}},
\bauthor{\binits{N.}~\bsnm{Parmar}},
\bauthor{\binits{J.}~\bsnm{Uszkoreit}},
\bauthor{\binits{L.}~\bsnm{Jones}},
\bauthor{\binits{A.N.}~\bsnm{Gomez}},
\bauthor{\binits{{\L}.}~\bsnm{Kaiser}} and
\bauthor{\binits{I.}~\bsnm{Polosukhin}},
\bctitle{Attention is all you need},
in: \bbtitle{Advances in neural information processing systems},
\byear{2017},
pp.~\bfpage{5998}--\blpage{6008}.
\end{bchapter}
\endbibitem

\bibitem[\protect\citeauthoryear{Wang et~al.}{2008}]{WBH08}
\begin{bchapter}
\bauthor{\binits{C.}~\bsnm{Wang}},
\bauthor{\binits{D.M.}~\bsnm{Blei}} and
\bauthor{\binits{D.}~\bsnm{Heckerman}},
\bctitle{Continuous Time Dynamic Topic Models},
in: \bbtitle{Proceedings of the $24^{\text{th}}$ Conference on Uncertainty in
  Artificial Intelligence},
\byear{2008}.
\end{bchapter}
\endbibitem

\bibitem[\protect\citeauthoryear{Wang and Mc{C}allum}{2006}]{WC06}
\begin{bchapter}
\bauthor{\binits{X.}~\bsnm{Wang}} and
\bauthor{\binits{A.}~\bsnm{Mc{C}allum}},
\bctitle{Topics over {t}ime: a non-{M}arkov continuous-time model of topical
  trends},
in: \bbtitle{Proceedings of the $12^{\text{th}}$ ACM SIGKDD International
  Conference on Knowledge Discovery and Data Mining},
\bpublisher{ACM},
\blocation{New York, NY, USA},
\byear{2006},
pp.~\bfpage{424}--\blpage{433}.
ISBN \bisbn{1-59593-339-5}.
doi:\doiurl{10.1145/1150402.1150450}.
\end{bchapter}
\endbibitem

\bibitem[\protect\citeauthoryear{Weston et~al.}{2011}]{weston2011wsabie}
\begin{bchapter}
\bauthor{\binits{J.}~\bsnm{Weston}},
\bauthor{\binits{S.}~\bsnm{Bengio}} and
\bauthor{\binits{N.}~\bsnm{Usunier}},
\bctitle{WSABIE: Scaling Up to Large Vocabulary Image Annotation.},
in: \bbtitle{IJCAI},
\beditor{\binits{T.}~\bsnm{Walsh}}, ed.,
\bpublisher{IJCAI/AAAI},
\byear{2011},
pp.~\bfpage{2764}--\blpage{2770}.
ISBN \bisbn{978-1-57735-516-8}.
\url{http://dblp.uni-trier.de/db/conf/ijcai/ijcai2011.html\#WestonBU11}.
\end{bchapter}
\endbibitem
\end{thebibliography}

\end{document}